\documentclass[acmconf,manuscript,screen,natbib=true]{acmart}

\usepackage{xcolor}
\usepackage[inline,nomargin,index]{fixme}

\fxsetup{theme=color,mode=multiuser,inlineface=\itshape,envface=\itshape,status=draft}
\FXRegisterAuthor{lk}{alk}{\colorbox{gray!10!white}{\color{black}Lizzie}}
\FXRegisterAuthor{as}{aas}{\colorbox{gray!10!white}{\color{black}Andrew}}
\FXRegisterAuthor{ah}{aah}{\colorbox{gray!10!white}{\color{black}Aaron}}
\FXRegisterAuthor{dr}{adr}{\colorbox{gray!10!white}{\color{black}Deb}}
\fxusetargetlayout{color}

\usepackage[threshold=1]{csquotes}

\newcommand{\ignore}[2]{\hspace{0in}#2}

\AtBeginDocument{%
  \providecommand\BibTeX{{%
    \normalfont B\kern-0.5em{\scshape i\kern-0.25em b}\kern-0.8em\TeX}}}

\copyrightyear{2022}
\acmYear{2022}
\setcopyright{acmlicensed}\acmConference[FAccT '22]{2022 ACM Conference on Fairness, Accountability, and Transparency}{June 21--24, 2022}{Seoul, Republic of Korea}
\acmBooktitle{2022 ACM Conference on Fairness, Accountability, and Transparency (FAccT '22), June 21--24, 2022, Seoul, Republic of Korea}
\acmPrice{15.00}
\acmDOI{10.1145/3531146.3533158}
\acmISBN{978-1-4503-9352-2/22/06}

\begin{document}

%\drerror{TO DO: references}
%\title{Dealing With Dysfunction in AI}
%\title{Algorithms are the New Snake Oil}
\title{The Fallacy of AI Functionality}
%: Dealing with Dysfunction in Deployed AI Systems
%The Fallacy of AI Functionality 
% The AI Functionality Functionality 

%%Algorithms are the New Snake Oil : Policy Responses to Dysfunctional AI 
%The emperor has no brain: Policy Responses to Dysfunctional AI 
%The Designed-In Dangers of Dysfunctional AI
%Dealing With Dysfunctional AI 
%Dealing With Dysfunction in AI 

\author{Inioluwa Deborah Raji*}
\affiliation{%
  \institution{University of California, Berkeley}
  \city{Berkeley}
  \state{CA}
  \country{USA}}
\email{rajiinio@berkeley.edu}

\author{I. Elizabeth Kumar*}
\affiliation{%
  \institution{Brown University}
  \city{Providence}
  \state{RI}
  \country{USA}}
\email{iekumar@brown.edu}

\author{Aaron Horowitz}
\affiliation{%
  \institution{American Civil Liberties Union}
  \city{New York City}
  \state{NY}
  \country{USA}}
\email{ahorowitz@aclu.org}

\author{Andrew D. Selbst}
\affiliation{%
  \institution{University of California, Los Angeles}
  \city{Los Angeles}
  \state{CA}
  \country{USA}}
\email{aselbst@law.ucla.edu }

\begin{abstract}

%For too long, policymakers have ignored the fact that many deployed AI systems do not work. They are constructed haphazardly, deployed indiscriminately, and promoted deceptively. AI policy cannot start with the assumption of functional behavior - there needs to be some accountability of performance claims being made by product vendors or institutional users. In this paper, we demonstrate through case studies in the US legal context how relying on corporate, academic and media narratives of functionality breaks AI accountability mechanisms in important ways - letting dysfunctional but “ethical” or value-aligned deployments through the cracks - while also leaving a lot of potential legal and organizational levers on the table. We present the assessment of a range of functionality failures as a necessary “first step” towards protecting the impacted. 

%Although an incomplete solution, we note that it provides the necessary initial friction that allows for further inquiry into other ethical issues.

%As the discourse around AI accountability has matured and  policymakers have moved into action, one aspect has so far been underappreciated: that many deployed AI systems do not work. 

Deployed AI systems often do not work. They can be constructed haphazardly, deployed indiscriminately, and promoted deceptively.
However, despite this reality, scholars, the press, and policymakers pay too little attention to functionality. This leads to technical and policy solutions focused on “ethical” or value-aligned deployments, often skipping over the prior question of whether a given system functions, or provides any benefits at all. %In this paper, we demonstrate through case studies in the US legal context how relying on corporate, academic and media narratives of functionality breaks AI accountability mechanisms in important ways, and
To describe the harms of various types of functionality failures, we analyze a set of case studies to create a taxonomy of known AI functionality issues. 
We then point to policy and organizational responses that are often overlooked and become more readily available once functionality is drawn into focus. We argue that functionality is a meaningful AI policy challenge, operating as a necessary first step towards protecting affected communities from algorithmic harm.

\end{abstract}




%Alt: 
%The emperor has no brain: Policy Responses to Dysfunctional AI 
%% -> How can we address AI's impact when the basic premise of the industry's claim turn out to be false? 
%The Designed-In Dangers of Dysfunctional AI
%Dealing With Dysfunctional AI 

\maketitle

\section{Introduction}

%\drerror{TO DO: Make it sound more concise} %DONE

%all these other industries have death with the snake oil problem and now it's AI's turn 

As one of over 20,000 cases falsely flagged for unemployment benefit fraud by Michigan’s MIDAS algorithm~\cite{charette2018michigan}, Brian Russell had to file for bankruptcy, undermining his ability to provide for his two young children. The state finally cleared him of the false charges two years later~\cite{bankrupt_MIDAS}. RealPage, one of several automated tenant screening tools producing “cheap and fast—but not necessarily accurate—reports for an estimated nine out of 10 landlords across the country”, flagged Davone Jackson with a false arrest record, pushing him out of low income housing and into a small motel room with his 9-year-old daughter for nearly a year~\cite{kirchner2020automated,kirchner2020access}. Josiah Elleston-Burrell had his post-secondary admissions potentially revoked~\cite{alevels,kippin2021covid}, Robert Williams was wrongfully arrested for a false facial recognition match~\cite{hill2020wrongfully}, Tammy Dobbs lost critical access to healthcare benefits~\cite{lecher2018happens}. The repercussions of AI-related functionality failures in high stakes scenarios cannot be overstated, and the impact reverberates in real lives for weeks, months and even years. %lives of those impacted 

Despite the current public fervor over the great potential of AI,
%corporations, the media, policymakers and certain critics---
many deployed algorithmic products do not work.
%Despite the insistence of corporations, the media, policymakers and certain critics---many deployed algorithmic products simply do not work. %In fact, it can seem as though everything, everywhere is falling apart.
AI-enabled moderation tools regularly flag safe content~\cite{tiktokerror, fb_hoes, fb_nudity}, teacher assessment tools mark star instructors to be fired~\cite{richardson2019litigating, paige2020houston}, 
%https://i-d.vice.com/en_uk/article/m7epya/tiktoks-algorithm-reportedly-bans-creators-using-terms-black-and-blm
%https://nypost.com/2021/07/20/facebook-cracks-down-on-discussing-hoes-in-gardening-group/
%https://www.businessinsider.com/facebook-mistakes-onions-for-sexualised-content-2020-10
%[Do we know what Danish vs. AAVE on Twitter referes to?]
hospital bed assignment algorithms prioritize healthy over sick patients~\cite{obermeyer2019dissecting}, and  medical insurance service distribution and pricing systems gatekeep necessary care-taking resources~\cite{lecher2018happens, richardson2019litigating}. Deployed AI-enabled clinical support tools 
%https://www.theverge.com/2018/3/21/17144260/healthcare-medicaid-algorithm-arkansas-cerebral-palsy 
% + litigating algos 
misallocate prescriptions~\cite{pain_wired}, 
%https://www.wired.com/story/opioid-drug-addiction-algorithm-chronic-pain/
misread medical images~\cite{freeman2021use,oakden2020hidden}, 
%https://www.bmj.com/content/374/bmj.n1872 ; Lauren Or paper 
and misdiagnose~\cite{Strickland_undated-ng, sepsis_validation}. The New York MTA's pilot of facial recognition had a reported 100\% error rate, yet the program moved forward anyway~\cite{berger2019mta}.
% https://www.wsj.com/articles/mtas-initial-foray-into-facial-recognition-at-high-speed-is-a-bust-11554642000}
%cite: IBM overpromised + sepsis 
%These failures reveal a broader pattern of a market saturated with dysfunctional, deployed AI products. %that don't work.%functionality issues that continue to be unaddressed.
Some of these failures have already proven to disproportionately impact some more than others: moderation tool glitches target minoritized groups~\cite{diaz2021double}; facial recognition tools fail on darker skinned female faces~\cite{Buolamwini_undated-dd}; a hospital resource allocation algorithm's misjudgements will mostly impact Black and lower income patients~\cite{obermeyer2019dissecting}. However, all failures in sum reveal a broader pattern of a market saturated with dysfunctional, deployed AI products.

Importantly, the hype is not limited to AI's boosters in corporations and the technology press; scholars and policymakers often assume functionality while discussing the dangers of algorithmic systems as well. %Despite this reality, m
In fact, many of the current critiques, policy positions and interventions in algorithmic accountability implicitly begin from the  premise that such deployed algorithmic systems work, echoing narratives of super-human ability~\cite{firestone2020performance},
%translation
broad applicability~\cite{grover},  
%grover 
and consistency~\cite{pineau2021improving}, 
%reproduciblity paper, 
espoused in corporate marketing materials, academic research papers and in mainstream media. These proposals thus often fall short of acknowledging the functionality issues in AI deployments and the role of the lack of functional safety in contributing to the harm perpetuated by these systems.
%Depending on the context of deployment, the result is often catastrophic harm. 
%Unemployment fraud detection tools lead to thousands of false accusations [cite] and in at least one case, a declaration of bankruptcy[cite]. Tenant screening errors lead to unjustly rejected housing applications[cite], incorrect grade adjustments block admissions [cite], and false identity matches lead to wrongful arrests [cite].
%TO add: criminal database and glitches? 

%However, w
 %How is the market saturated 
%AI-advertised hiring tools lack simple reporting of any validation processes [cite]. Deployed sepsis models demonstrate no testing of external validity [cite].
%https://jamanetwork.com/journals/jamainternalmedicine/article-abstract/2781313
%The unemployment fraud detection algorithm was inappropriately evaluated [cite, cite] and performance of facial recognition tools deceptively presented [cite]. 
%TO DO: find example 
%\drnote{combo para}
\ignore{The myth of functionality is one held dearly by corporate stakeholders and their investors.} If a product works, we can weigh its costs and benefits. But if the product does \emph{not} work, the judgment is no longer a matter of pros and cons, but a much simpler calculation, exposing that this product does not deserve its spot on the market. 
%This is the myth that corporate stakeholders and their investors hold most dearly: if a product is revealed to not work, the judgement of its value is no longer a consideration of pros and cons or dual use, but a much simpler calculation, which exposes that a product does not deserve its spot on the market.
%\drnote{part of this is repeated from previous functionality definition->}
Although notions of accuracy and product expectations are stakeholder-dependent and can be contested, the assessment of such claims are often easier to empirically measure, grounding the discussion of harm in a way that is challenging to repudiate.

As an overlooked aspect of AI policy, functionality is often presented as a consideration secondary to other ethical challenges. In this paper, we argue that it is a primary concern that often precedes such problems.
%, and should likely be dealt with first. 
We start by calling out what we perceive to be a functionality assumption, prevalent in much of the discourse on AI risks. We then %directly challenge the feasibility of 
argue that this assumption does not hold in a large set of cases. %, demonstrating the many ways in which AI deployment failures interfere with narratives that  %that fails to consider  
Drawing on the AI, Algorithmic and Automation Incident and Controversy Repository (AAAIRC), we offer a taxonomy of the ways in which such failures can take form and the harms they cause, which differ from the more commonly cited critiques of AI.
%To demonstrate functionality harms, we look at incidents of harmful dysfunction of \emph{deployed} products on the market. Drawing on the AI, Algorithmic and Automation Incident and Controversy Repository (AAAIRC), we offer a taxonomy of failures can take form and the harms they cause, which differ from the more commonly cited critiques of AI. 
We then discuss the existing accountability tools to address functionality issues, that are often overlooked in AI policy literature and in practice, due in large part to this assumption of functionality.

\section{Related Work}

%Past conversations about 
%In this section, we outline some brief background and related work. We start with a discussion of criti-hype and the tendency for even those that oppose AI deployment to fall for the myth of functionality. 
%We then mention some relevant historical context, pointing to other industries that have faced a similar crisis and those already drawing parallels to the AI industry. We end by discussing the state of past discussions of AI's functionality issues in deployment, and note a set of valid policy observations related to this challenge.
%Overall, a 
A review of past work demonstrates that although there is some acknowledgement that AI has a functionality problem, little has been done to systematically discuss the range of problems specifically associated with functionality.% in AI deployment.%, and more importantly there is a lack of systematic mapping from the challenge to discourse in algorithmic accountability. While many of the common critiques of AI can be categorized as AI "not working" in the sense of misaligning with important values or causing harm, few address the basic functionality requirement head on. In fact, as we describe in Section \ref{assumption}, many of those critiques actually \textit{assume} basic functionality as a premise.  %the lack of this discourse in discussions of algorithmic accountability.  
Recent work details 
%extensive evidence that the AI field, and specifically machine learning research as a whole, 
that the AI research field suffers from scientific validity and evaluation problems~\cite{haibe_kains, mit_replication}.
\citet{kapoor_irreproducible_2021} have demonstrated 
%specific instances of 
reproducibility failures in published work on predicting civil wars. 
\citet{LiaoAreWe2021} found that 
%due to internal and external validity issues, 
advances in machine learning often ``evaporate under closer scrutiny or turn out to be less widely applicable than originally hoped.''
%In particular, inadequate evaluation design can distort understanding of how a model may behave in practice---with static benchmarks often representing scoped and static representations of the deployment environment in ways that can be easily manipulated or incomplete~\cite{grover}. For example, benchmark datasets sourced from biased populations will easily disguise performance failures on underrepresented subgroups~\cite{barocas2021designing,raji2020saving}. 

There is also some work demonstrating that AI products are challenging to engineer correctly in practice.
In a survey of practitioners, \citet{ml_software_practices} describe how developers often modify traditional software engineering practices due to unique challenges presented by ML, such as the increased effort required for testing and defining requirements. They also found that ML practitioners "tend to communicate less frequently with clients" and struggle to make accurate plans for the tasks required in the development process.
\citet{sculley2015hidden} have additionally argued that ML systems "have a special capacity for incurring technical debt."

Other papers discuss how the AI label lends itself to inflated claims of functionality that the systems cannot meet. \citet{fake_ai} and \citet{broussard2018artificial} critique hyped narratives pushed in the AI industry, joined by many similar domain-specific critiques~\cite{tennant2021attachments,bender2021dangers,stark_and_hutson,sloane2022silicon,raghavan2020mitigating,bender2020climbing}. 
%such as  \citet{tennant2021attachments}'s critique of autonomy rhetoric in self-driving cars and \citet{bender2021dangers}'s discussion of the limitations of %language understanding capabilities of large language models. 
\citet{narayanan2019recognize} recently popularized the metaphor of ``snake oil'' as a description of such AI products, raising concerns about the hyperbolic claims now common on the market today.
%noting the similarities between the ancient snake oil industry and the
\citet{richardson2021defining} has noted that despite the "intelligent" label, many deployed AI systems used by public agencies involve simple models defined by manually crafted heuristics. Similarly, \citet{grover} argue that AI makes claims to generality while modeling behaviour that is determined by highly constrained and context-specific data. In a study of actual AI policy discussions, \citet{krafft_et_al} found that policymakers often define AI with respect to how human-like a system is, and concluded that this could lead to deprioritizing issues more grounded in reality. %and questions of performance.  %https://arxiv.org/abs/1912.11095 

%These discussion do not center the problem of functionality as a general point, and thus do not discus
%rather than exploring how these model failures play out in AI-related products, and noting the policy implications of those challenges upon release in the market.  
%\lknote{Any other useful stuff in Fake AI book?, Artificial Unintelligence, etc}
%\drnote{ AU: technochauvinism -> ; people talk about imagined technologies vs. the reality of how these systems}
%\asnote{I would just list them as calling out AI not working, perhaps?.}

Finally, \citet{vinsel_critihype} has argued that even critics of technology often hype the very technologies that they critique, as a way of inflating the perception of their dangers. %describing them as more dangerous. 
He refers to this phenomenon as "criti-hype"---criticism which both needs and feeds on hype. As an example, he points to disinformation researchers, who embrace corporate talking points of a recommendation model that can meaningfully influence consumer behavior to the point of controlling their purchases or voting activity---when in actuality, these algorithms have little ability to do either~\cite{hwang2020subprime, badnews, robertson2021engagement,gibney2018scant,hern2018cambridge}. Even the infamous Cambridge Analytica product was revealed to be ``barely better than chance at applying the right [personality] scores to individuals'', and the company accused explicitly of ``selling snake oil''~\cite{hern2018cambridge}.  %Similar arguments have been made about the Cambridge Analytica scandal---the outrage about how their psychographic profiles influencing the 2016 election based a questionable assumption that their product worked in the first place~\cite{gibney2018scant}. In a testimony to the UK parliment, lead researcher Dr. Aleksandr Kogan goes so far as to suggest that ``it was barely better than chance at applying the right [personality] scores to individuals'' and that ``maybe [Cambridge Analytica] was selling snake oil''~\cite{hern2018cambridge}. 

\section{The Functionality Assumption}
\label{assumption}

It is unsurprising that promoters of AI
%---often corporate actors or media, as well as some academics--
do not tend to question its functionality.
% Instead, they are prone to applying the AI label as broadly as possible and overclaiming its capabilities, relying on imagined technologies to sell the future.
%This phenomenon of hype has been discussed at length elsewhere~\cite{slota2020good}.
More surprising is the prevalence of criti-hype in the scholarship and political narratives around automation and machine learning---even amidst discussion of valid concerns such as trustworthiness, democratization, fairness, interpretability, and safety. These fears, though legitimate, are often premature “wishful worries”---fears that can only be realized once the technology works, or works "too well", rather than being grounded in a reality where these systems do not always function as expected ~\cite{vinsel_critihype}.
In this section, we discuss how criti-hype in AI  manifests as an unspoken assumption of functionality. %This unchallenged belief that AI works as advertised impacts media, academic and corporate narratives of how well AI systems work and deeply influences the AI policy proposals we see today. 
%We first present practical examples of the false narrative of functionality promoted by media, academia and industry, and then introduce the ways in which these narratives coalesce into the unquestioning acceptance of AI product function in policy developments. 
%We present functionality as a valid harm to be considered as a preliminary system requirement, before we can appropriately address the slew of other ethical concerns plaguing the AI industry, but can observe several instances of functionality being overlooked as a concern when addressing several policy developments in the AI context. 
%We thus see several instances of functionality being overlooked as a concern when addressing several policy developments in the AI context. 

% trust
%\subsection{Trust}

The functionality of AI systems is  % so much taken for a given that it is
rarely explicitly mentioned in AI principle statements, policy proposals and AI ethics guidelines. In a recent review of the landscape of AI ethics guidelines, \citet{Jobin2019-oa} found that few acknowledge the possibility of AI not working as advertised.
% Even descriptions of supposedly related concepts such as non-malfeasance and trust still hold this assumption of technology that works as intended.
% In the case of malfeasance guidelines, the concern is articulated to be mainly that of malicious use of supposedly functional AI products by nefarious actors.
In guidelines about preventing malfeasance, the primary concern is malicious use of supposedly functional AI products by nefarious actors.
% For trust, the cited documents are more geared towards eliciting trust from users or the public of AI products. Trust in these products is assumed to be to the benefit of these stakeholders, in order for AI to ``fulfill its world changing potential’’~\cite{} -- indicating an assumption inherent to these proposals of functional, and generally beneficial AI products.
Guidelines around "trust" are geared towards eliciting trust in AI systems from users or the public, implying that trusting these AI products would be to the benefit of these stakeholders and allow AI to ``fulfill its world changing potential’’~\cite{Jobin2019-oa}.
Just one guideline of the hundreds reviewed in the survey ``explicitly suggests that, instead of demanding understandability, it should be ensured that AI fulfills public expectations’’~\cite{Jobin2019-oa}.
%\lknote{Context for who NIST is and in what context they are saying this?}\drnote{Done} 
Similarly, the U.S. National Institute of Standards and Technology (NIST) seeks to define "trustworthiness" based primarily on how much people are willing to use the AI systems they are interacting with~\cite{Stanton2021-oa}. This framing puts the onus on people to trust in systems, and not on institutions to make their systems reliably operational, in order to earn that trust%worthy% by living up to performance and reliability expectations
~\cite{aclu-comment-trust, noonetrustai-xo}. NIST's concept of trust is also limited, citing the "dependability" section of ISO/IEEE/IEC standards~\cite{ieee_dictionary_dependability}, but leaving out other critical concepts in these dependability engineering standards that represent basic functionality requirements, including assurance, claim veracity, integrity level, systematic failure, or dangerous condition. 
%"assurance", "claim veracity, "integrity level",  "systematic failure" or "dangerous condition". 
Similarly, the international trade group, the Organisation for Economic Co-operation and Development (OECD), mentions "robustness” and “trustworthy AI” in their AI principles but makes no explicit mention of expectations around basic functionality or performance assessment~\cite{yeung2020recommendation}.% The focus is instead on “the ability to withstand or overcome adverse conditions, including digital security risks”, broader “trace-ability” and “risk assessment”~\cite{yeung2020recommendation}. %\lknote{who is OECD} \drnote{addressed}

%democratization
%\subsection{Democratization}

The ideal of “democratizing” AI systems, and the resulting AI innovation policy, is another effort premised on the assumed functionality of AI. This is the argument that access to AI tooling and AI skills should be expanded %\cite{democratizing_cloud, democratizing_h20, democratizing_deloitte}
\cite{democratization, democratization2, democratization3, democratizing_cloud}---with the corollary claim that it is problematic that only certain institutions, nations, or individuals have access to the ability to build these systems~\cite{de_democratizing}.
% These democratization efforts at an international scale are often tied to existing geo-political dynamics. 
%With the onset of COVID-19 tools, much of the US policy response~\cite{us-covid}, which included calls for funding and data access opportunities~\cite{covid-ostp}, was spurred by similar activity in China~\cite{c-covid-1, c-covid-2}, Greece~\cite{g-covid} and elsewhere. Efforts to implement AI tools in the Global South were sidelined due to the lack of resources and data~\cite{covidai-africa}, and the global digital policy focused more and more on the concept of increasing the range of participants that could access information and develop AI systems to address the pandemic crisis. 
A recent example of democratization efforts was the global push for the relaxation of oversight in data sharing in order to allow for more innovation in AI tool development in the wake of the COVID-19 pandemic~\cite{covid_intl, covid_us, covid_greece, covid_china, covid_africa}. % \drerror{TO DO: citations}.
The goal of such efforts was to empower a wider range of non-AI domain experts to participate in AI tool development. This policy impact was long lasting and informed later efforts such as the AI National Resource (AINR) effort in the US~\cite{NAIR} and the National Medical Imaging Platform (NMIP) executed by National Health Services (NHS) in the UK~\cite{NMIP}. 
%Thus,  there was an influx of AI projects deployed and products introduced to the market~\cite{covid-AI}, driven by AI policy developments centered on democratization~\cite{OECD-covid, digi-policy}, with international efforts being launched and debated~\cite{council} on how to make data and computation more easily available in order to drive innovation in the development of AI systems. 
In this flurry of expedited activity, some parallel concerns were also raised 
%by policymakers, attempting to pre-emptively address a range of issues thought to arise in consequence of ``democratization’’ efforts -- including 
about how the new COVID-19 AI tools would adequately address cybersecurity, privacy, and anti-discrimination challenges~\cite{krass2021us, digiwatch}, but the functionality and utility of the systems remained untested for some time~\cite{covidfail_summ, covidfail1, covidfail2, covidfail3}. 
%However, years later, it became clear that, according to several retrospective studies, many of the sponsored and public-data supported tools meant to solve problems in the global crisis, did nothing but create new problems\cite{MITTR-covid}. Put simply - these tools did not work, with several of these innovation policy and democratization efforts yielding technology that wound up being unsuitable for clinical deployment~\cite{medical-covidfail, turing-covidfail}. In fact, according to one study, of the 415 AI tools analyzed, none were fit for clinical deployment~\cite{survey-covidfail}. In the rush to push more products into the market under the name of democratisation, there was an influx of AI tools released prematurely, that could not live up to expectations of performance and ended up being completely meaningless. 

%”safety”,  AGI/ "AI is too smart"/safety + "ethical"/moral machine
%\subsection{Safety}

An extremely premature set of concerns are those of an autonomous agent becoming so intelligent that humans lose control of the system. While it is not controversial to claim that such concerns are far from being realized %\lkerror{citations!! (Deb?)} \drnote{could only think of the infamous op-ed}
\cite{crawford2016artificial, atkinson2018going, prunkl2020beyond}, this fear of misspecified objectives, runaway feedback loops, and AI alignment presumes the existence of an industry that can get AI systems to execute on any clearly declared objectives, and that the main challenge is to choose and design an appropriate goal. %the choice and design of an appropriate objective. %\lknote{Ok- I really wanted to bring back my "paperclip" discussion that was the rest of this paragraph so I did but we can talk about taking it out again later lol} \drerror{ok. TO DO: re-write}
%This is best illustrated by the infamous ``paperclip'' thought experiment, which describes an out-of-control AI system destroying all of humanity to optimize on its objective of producing as many paperclips as possible -- a scenario taken seriously as a concern by major scholars and research labs~\cite{bostrom2014superintelligence, brundage2015taking}. 
Needless to say, if one thinks the danger of AI is that it will work too well~\cite{shane2019janelle}, it is a necessary precondition that it works at all.

The fear of hyper-competent AI systems also drives discussions on potential misuse~\cite{brundage2018malicious}. For example, expressed concerns around large language models %is also prone to subscribing to the assumption of functionality.
%Many with ethical concerns surrounding large language models
centers on hyped narratives of the models' ability to generate hyper-realistic online content, which could theoretically be used by malicious actors to facilitate harmful misinformation campaigns~\cite{llmdeepmind, llmopenai}. 
While these are credible threats, concerns around large language models tend to dismiss the practical limitations of what these models can achieve~\cite{bender2021dangers}, %and ethical discussions focus disproportionately on regulating the malicious generation of fake content~\cite{llmdeepmind, llmopenai} rather than 
neglecting to address more mundane hazards tied to the premature deployment of a system that does not work~\cite{ettinger2020bert, goog_search_fail}. 
This pattern is evident in the EU draft AI regulation~\cite{AI_Act}, where, even as the legislation does concern functionality to a degree, the primary concerns---questions of ``manipulative systems,'' ``social scoring,'' and ``emotional or biometric categorization''---``border on the fantastical''
% are prohibited or restricted due to moral reasoning around potential malicious use, with little mention of concerns for dysfunction, false claims or the impossibility of the modeling tasks
~\cite[p.~98]{veale_euact}. %\drnote{I want to specify p.11 in the cite}. 
A major policy focus in recent years has been addressing issues of bias and fairness in AI. Fairness research is often centered around attempting to balance some notion of accuracy with some notion of fairness~\cite{friedler2019comparative, fish2016confidence, disparate_impact_suresh}. This research question presumes that an unconstrained solution without fairness restrictions is the optimal solution to the problem. However, this intuition is only valid when certain conditions and assumptions are met~\cite{Mitchell2021-pk, impossibility_of_fairness, fairness_tradeoffs_neurips}, such as the measurement validity of the data and labels. Scholarship on fairness also sometimes presumes that unconstrained models will be optimal or at least useful. \citet[p.~707]{barocas2016big} argued that U.S. anti-discrimination law would have difficulty addressing algorithmic bias because the "nature of data mining" means that in many cases we can assume the decision is at least statistically valid. 
%The arguments by  \citet{barocas2016big} split between cases between those of AI tools introducing new biases into a system (due to problems with the data, or the interaction between the target variable or class labels and the real world), and cases where AI picks up on latent biases in the real world, as remnants of historical discrimination. While they refer to the former cases as "imperfect" data mining (as opposed to the latter's hypothetical "perfect case of data mining"), the critique assumes basic functionality. For example, in their legal analysis, they observe: "Once a target variable is established as job related, the first question is whether the model is predictive of that trait. The nature of data mining suggests that this will be the case."~\cite[p.~707]{barocas2016big}.
Similarly, as an early example of technical fairness solutions, \citet{feldman2015certifying} created a method to remove disparate impact from a model while preserving rank, which only makes sense if the unconstrained system output is correct in the first place. %These are just two examples of many.
%\drnote{+ something about a Suresh paper on 4/5th rule mentioned in the meeting?}
Industry practitioners then carry this assumption into how they approach fairness in AI deployments. For example, audits of AI hiring tools focus primarily on ensuring an 80\% selection rate for protected classes (the so-called 4/5ths rule) is satisfied, and rarely mention product validation processes, demonstrating an assumed validity of the prediction task~\cite{raghavan2020mitigating, wilson2021building, engler2021independent}.

Another dominant theme in AI policy developments is that of explainability or interpretability. %For example, this was thought to be a major consideration for adequate ML and AI deployment in healthcare by relevant regulators such as the FDA ~\cite{shaban2021explainability, missing_link_xai}.
%This policy consideration, included in Europe’s General Data Protection Regulation (GDPR)~\cite[Arts.~13--15, 22]{GDPR} and addressed as early as 2016 by Defense Advanced Research Projects Agency (DARPA), arose as an attempt to open up the ``black box’’ of AI models, making internal algorithmic decision-making processes more visible ~\cite{adadi2018peeking}. 
The purpose of making models explainable or interpretable differs depending on who is seen as needing to understand them. From the engineering side, interpretability is
%actually focused directly on functionality. Most of the reason that interpretability is
usually desired 
%in technical work is 
for debugging purposes~\cite{bhatt2020explainable}, so it is focused on functionality. But on the legal or ethical side, things look different. There has been much discussion about whether the GDPR includes a "right to explanation" and what such a right entails~\cite{kaminski2019right, selbst2017meaningful, wachter2017right, edwards2017slave}. Those rights would serve different purposes. To the extent the purpose of explanation is to enable contestation~\cite{kaminski2021right}, then functionality is likely included as an aspect of the system subject to challenge. To the extent explanation is desired to educate consumers about how to improve their chances in the future~\cite{barocas2020hidden}, such rights are only useful when the underlying model is functional.
%and, if automatically generated, the explanation-generating model are functional. 
Similarly, to the extent regulators are looking into functionality, %---which as we argue here, might be minimal---
explanations aimed at regulators can assist oversight, but typically explanations are desired to % it is considered necessary to explain %to regulators what the model is doing and 
check the basis for decisions, while assuming the systems work as intended.

Not all recent policy developments hold the functionality assumption strongly. %A bill proposed in Congress in 2019---the Algorithmic Accountability Act of 2019---made a passive acknowledgement of the possibility of dysfunctional 
%\lknote{ok we need a better word than unideal but idk what you were going for}
%\drnote{hm - imperfect?}
%AI systems, referring to ``risks that the automated decision system may result in or contribute to inaccurate, unfair, biased, or discriminatory decisions impacting consumers’’~\cite{AAA}.
%\drnote{not sure how to cite the Act itself}. 
The Food and Drug Administration (FDA) guidelines for AI systems integrated into software as a medical device (SaMD) has a strong emphasis on functional performance, clearly not taking product performance as a given~\cite{FDA}. The draft AI Act in the EU includes requirements for pre-marketing controls to establish products' safety and performance, as well as quality management for high risk systems~\cite{veale_euact}.
%\drnote{Add the new version of the EU Act addressing functionality as a positive example here?} 
%\drnote{The EU AI Act does make mention of ``pre-market'' vetting.}
These mentions suggest that functionality is not always ignored outright. Sometimes, it is considered in policy, but in many cases, that consideration lacks the emphasis of the other concerns presented.

\section{The Many Dimensions of AI Dysfunction}
\label{taxonomy}

%\drnote{TO DO: re-visit definition}
%\drnote{operationally falling short of expectations of stakeholders  -- expanding the definition of stakeholders beyond those that develop the systems and the claims they make, but also including the public, government, scientific validity, etc.}

Functionality can be difficult to define precisely. The dictionary definition of ``fitness for a product's intended use''~\cite{OED} is useful, but incomplete, as some intended uses are impossible. Functionality could also be seen as a statement that a product lives up to the vendor's performance claims, but this, too, is incomplete; specifications chosen by the vendor could be insufficient to solve the problem at hand. 
%expectations of other stakeholders beyond the vendor should be considered. %Ultimately, at the boundaries, functionality is still a judgment call, albeit a more concrete one than many. 
%We anchor our understanding of functionality to the notion of operationally meeting the articulated performance and system behavior expectations of a broad set of involved stakeholders. %ranging from the vendors themselves to the public, the scientific community, government or others.
%with respect to performance or operation, rather than ethical concerns.
Another possible definition is "meeting stakeholder expectations" more generally, but this is too broad as it sweeps in wider AI ethics concerns with those of performance or operation.

Lacking a perfectly precise definition of functionality, in this section we invert the question by creating a taxonomy that brings together disparate notions of product failure. %to define the term with a taxonomy we provide. 
%To flesh out the idea of functionality failures, in this section, we present a taxonomy of different ways in which AI systems fail, pointing to specific instances where the functionality assumption was demonstrably false in consequential situations.  
% We taxonomize these failures to illustrate the scope of the issue, and offer language in which to ground future discussions of functionality in research and policy. Throughout, we define the absence of functionality to be any divergence between the realized behavior a system and the expectations for performance as articulated by the developer of the system, the public or other stakeholders such as regulators.
%the performance claims made by those that developed the system.
% \lkerror{Consistency- *IS* this exactly how we're defining it throughout? Feel like we should double check that. Also, I liked my 3-point explanation of our taxonomy, bringing it back here:}
Our taxonomy serves several other purposes, as well. Firstly, the sheer number of points of failure we were able to identify illustrates the scope of the problem. Secondly, we offer language in which to ground future discussions of functionality in research and policy. Finally, we hope that future proposals for interventions can use this framework to concretely illustrate the way any proposed interventions might work to prevent different kinds of failure. 
%\aserror{Something about why we don't define functionality}

\subsection{Methodology}

To challenge the functionality assumption and demonstrate the various ways in which AI doesn't work, we developed a taxonomy of known AI failures through the systematic review of case studies. %We first compiled examples of products marketed as AI failing in ways that perpetuated harm that appeared in the news or academic literature that was notable for failing at some tasks. 
To do this, we partly relied on the AI, Algorithmic and Automation Incident and Controversy Repository (AIAAIC) spreadsheet crowdsourced from journalism professionals~\cite{AIAAIC}. Out of a database of over 800 cases, we filtered the cases down to a spreadsheet of 283 cases from 2012 to 2021 based on whether the technology involved claimed to be AI, ML or data-driven, and whether the harm reported was due to a failure of the technology. In particular, we focused on describing the ways in which the artifact itself was connected to the failure, as opposed to infrastructural or environmental "meta" failures which caused harm through the artifact. We split up the rows in the resulting set and used an iterative tagging procedure to come up with categories that associate each example with a different element or cause of failure. We updated, merged, and grouped our tags in meetings between tagging sessions, resulting in the following taxonomy. We then chose known case studies from the media and academic literature to illustrate and best characterize these failure modes. %Details of our filtered spreadsheet can be found in the Appendix. 
\subsection{Failure Taxonomy}

Here, we present a taxonomy of AI system failures and provide examples of known instances of harm. 
%This section provides material evidence of how the functionality assumption is often incorrect, and dangerously breaks down in consequential contexts.
Many of these cases are direct refutations of the specific instances of the functionality assumptions in Section \ref{assumption}. 
%\lkerror{Todo: Update Table}

%\drnote{Functional safety}
%What It Means Not to “Work”? 
%(False advertising) Failing to adhere to performance claims made by the developer of the tool itself
%(Inappropriate use) Failing to be deployed within known functional limits (ie. not just shoddy builds but also limits to what you can even do with the technology)
%Definition :
%Wrong answer? It’s possible to get the right answer but you have the wrong answer. 
%Given some claim of performance/accuray, falling short 
%Given some notion of correctness, contradiction
%Falsifiable claims [we can say it’s wrong because there is some natural notion of correction]
%A couple other not to works to consider, may be overlapping/reducable:
%* (Impossible claims) Claims made about capabilities that AI can not do -- thinking of poor construct validity, phrenology stuff, etc.
%* (Expected Off-Brand Usage) Probably related to inappropriate use, but many validations will be done in such a way for the model to look good, even though intended users will surely not use it that way (eg mugshots,  companies setting "recommendation" of 66\% population threshold when its known that companies will use with applications where they are hiring 1% of a large pool)

% \drerror{TO DO (nice to have): expanded table with references from AI sucks spreadsheet}
\begin{table}[ht]
  \caption{Failure Taxonomy}
  \label{tab:taxonomy}
  \begin{tabular}{ll}
        \toprule
Impossible Tasks & Conceptually Impossible \\
    & Practically Impossible\\
    \midrule
    Engineering Failures & Design Failures \\
    %construct validity 
    & Implementation Failures \\
    % internal validity 
    & Missing Safety Features \\
    \midrule
    Post-Deployment Failures & Robustness Issues \\
    % robustness = external validity 
    & Failure under Adversarial Attacks \\
    % also include security risk 
    & Unanticipated Interactions \\
    \midrule
    Communication Failures & Falsified or Overstated Capabilities \\
    & Misrepresented Capabilities \\
    \bottomrule
  \end{tabular}
\end{table}

\subsubsection{Impossible Tasks}

In some situations, a system is not just "broken" in the sense that it needs to be fixed. Researchers across many fields have shown that certain prediction tasks cannot be solved with machine learning.
These are settings in which no specific AI developed for the task can ever possibly work, and a functionality-centered critique can be made with respect to the task more generally.
% is, from a certain perspective, the easiest to make, because if enough evidence already exists that a problem setting is fundamentally inappropriate, the study of how well any particular system works becomes redundant. On the other hand, 
Since these general critiques sometimes rely on philosophical, controversial, or morally contested grounds, the arguments can be difficult to leverage practically and may imply the need for further evidence of failure modes along the lines of our other categories.

%In some situations, a system is not just "broken" in the sense that it needs to be fixed. Researchers across many fields have shown that certain tasks cannot be solved with machine learning. These are the settings in which a functionality-based critique is, from a certain perspective, the easiest to make, because if enough evidence already exists that a problem setting is fundamentally inappropriate, the study of how well any particular system works becomes redundant. On the other hand, these "broad" critiques sometimes rely on philosophical, controversial, or morally contested grounds, which, as we suggested earlier, may imply the need for further evidence of failure modes along the lines of our other categories.

\paragraph{Conceptually Impossible}

Certain classes of tasks have been scientifically or philosophically "debunked" by extensive literature. In these cases, there is no plausible connection between observable data and the proposed target of the prediction task. This includes what Stark and Hutson call “physiognomic artificial intelligence,” which attempts to infer or create hierarchies about personal characteristics from data about their physical appearance~\cite{stark_and_hutson}. %\citet{veale_euact} specifically criticize the EU Act's failure to address this inconvenient truth for emotion detection tools.
Criticizing the EU Act's failure to address this inconvenient truth, \citet{veale_euact} pointed out that 
``those claiming to detect emotion use oversimplified, questionable taxonomies; incorrectly assume universality across cultures and contexts; and risk `[taking] us back to the phrenological past' of analysing character traits from facial structures.''

% \drnote{People are talking about it in policy as if its a real thing and its not}
%As mentioned in Section 3,  \cite{Veale,p. 11} observe  the assumed functionality of this technology in the EU AI Act, \drnote{cite EU Act section} %`` those claiming to detect emotion use oversimplified, questionable taxonomies; incorrectly assume universality across cultures and contexts; and risk ’[taking] us back to the phrenological past’ of analysing character traits from facial structures,’’ 
% making the provision insufficient in addressing the proliferation of a technology that claims capabilities that have been thoroughly debunked \cite{lukephrenology}. 

A notorious example of technology broken by definition are attempts to infer “criminality” from a person’s physical appearance. A paper claiming to do this “with no racial bias” was announced by researchers at Harrisburg University in 2020, prompting widespread criticism from the machine learning community~\cite{wired_criminality}. In an open letter, the Coalition for Critical Technology %claim that the task is impossible, 
note that the only plausible relationship between a person’s appearance and their propensity to commit a crime is via the biased nature of the category of “criminality” itself~\cite{coalition}. In this setting, %there is so much evidence supporting this claim that it should not be necessary to dig into what the researchers did or how they came up with their results: 
there is no logical basis with which to claim functionality.

\paragraph{Practically Impossible}
%Even if there is no conceptual argument against a type of prediction task being statistically possible,
There can be other, more practical reasons for why a machine learning model or algorithm cannot perform a certain task. For example, in the absence of any reasonable observable characteristics or accessible data to measure the model goals in question, attempts to represent these objectives end up being inappropriate proxies. As a construct validity issue, the constructs of the built model could not possibly meaningfully represent those relevant to the task at hand ~\cite{Jacobs2021-og,Jacobs2021-rk}.

\ignore{Criminal justice offers a wide variety of such practically impossible tasks, either through the lack of availability of the data.}Many predictive policing tools are arguably %presents a good example of 
practically impossible AI systems. Predictive policing attempts to predict crime at either the granularity of location or at an individual level~\cite{Ferguson2016-bs}. The data that would be required to do the task properly---accurate data about when and where crimes occur---does not and will never exist. While crime is a concept with a fairly fixed definition, it is practically impossible to predict because of structural problems in its collection. The problems with crime data are well-documented---whether in differential victim crime reporting rates~\cite{Akpinar2021-fb}, selection bias based on policing activities~\cite{Lum2016-hz,Ensign2017-vi}, dirty data from periods of recorded unlawful policing~\cite{Richardson2019-cn}, and more. 
%Predictive policing is a practically impossible task because t
%\drerror{I like this discussion on thick concepts -- wondering if it's worth it to revise a stream-lined version of this, but not tied to it}
%\drnote{In fact, many common criminal justice algorithm objectives such as "likelihood of flight","public safety", or "dangerousness" lack universal or legally valid constructs ~\cite{Mayson_dangdefendants,Slobogin2003-ou,Gouldin_undated-oc, Stevenson2021-fr}, and are thus typically determined by the value judgements of researchers~\cite{Alexandrova_undated-mx}.}
\ignore{
\asnote{What work is this discussion of thick concepts doing? Can we delete? Why is it in this subsection while it says they are conceptually impossible? Also, is it necessarily accurate that thick concept = conceptually impossible?}
Some of these are thick concepts -- concepts that describe and evaluate simultaneously -- which ideally are co-produced but are typically based on the values judgement of researchers~\cite{Alexandrova_undated-mx}. Common focus areas of criminal justice algorithms like "likelihood of flight","public safety", or "dangerousness" represent thick concepts. A universal or legally valid version of these constructs may not be possible~\cite{Mayson_dangdefendants,Slobogin2003-ou,Gouldin_undated-oc, Stevenson2021-fr}, making them  potentially \emph{conceptually} impossible. }

Due to upstream policy, data or societal choices, AI tasks can be practically impossible for one set of developers and not for another, or for different reasons in different contexts. The fragmentation, billing focus, and competing incentives of the US healthcare system have made multiple healthcare-related AI tasks practically impossible
% whereas in NHS slow shifts from paper to centralized electronic records have created different practical impossibility limits.
~\cite{Agrawal2020-rs}.
US EHR data is often erroneous, miscoded, fragmented, and incomplete~\cite{Hoffman2013-oa,Hoffman2013-ms}, creating a mismatch between available data and intended use~\cite{Gianfrancesco2018-vl}. Many of these challenges appeared when IBM attempted to support cancer diagnoses. In one instance, this meant using synthetic as opposed to real patients for oncology prediction data, leading to "unsafe and incorrect" recommendations for cancer treatments~\cite{Ross2018-nn}. In another, IBM worked with MD Anderson to work on leukemia patient records, poorly extracting reliable insights from time-dependent information like therapy timelines---the components of care most likely to be mixed up in fragmented doctors' notes~\cite{Strickland_undated-ng, Simon2019-ed}.

%Recently, IBM has found some success in narrower applications that may not share the practical impossibility problems of using EHR data for cancer diagnoses, like providing reports to oncologists from genetic information, though effects on patient outcomes remain unknown~\cite{Strickland_undated-ng}.  

%\drnote{Deb, if you can find something about law differences specifically creating the problem, happy to use that. The gold standard part of the paper was actually saying that Watson isnt allowed to be used bc its too good which ... I dont buy} \drerror{Really? Let me take another look}

%\ahnote{One area of practical impossibility I WISH we had space or time to cover is rare events prediction. Its a bit jumbled up in other sections for now (eg threshold level reco by Amazon Rekogntion) but I see this form of thinking a lot in public sector} \drnote{The rare prediction thing comes up in the IBM case as well}

%Disability monitoring and insurance 

\subsubsection{Engineering Failures}

Algorithm developers maintain enormous discretion over a host of decisions, and make choices throughout the model development lifecycle. These engineering choices include defining problem formulation~\cite{Passi2019-av}, setting up evaluation criteria~\cite{Passi2020-dr,LiaoAreWe2021}, and determining a variety of other details~\cite{Passi2018-jt,Muller2019-cy}. Failures in AI systems can often be traced to these specific policies or decisions in the development process of the system.

\paragraph{Model Design Failures}

Sometimes, the design specifications of a model are inappropriate for the task it is being developed for. 
%When specifying a machine learning model for a task, developers have many choices at their disposal. In a classification model, this includes 
For instance, in a classification model, choices such as which input and target variables to use, whether to prioritize accepting true positives or rejecting false negatives, and how to process the training data all factor into determining model outcomes. These choices are normative and may prioritize values such as efficiency over preventing harmful failures~\cite{Lehr_undated-aq, Dobbe2019-ms}.

%For example, i
In 2014, BBC Panorama uncovered evidence of international students systematically cheating on English language exams run by the UK's Educational Testing Service by having others take the exam for them. The Home Office began an investigation and campaign to cancel the visas of anyone who was found to have cheated. In 2015, ETS used voice recognition technology to identify this type of cheating. According to the National Audit Office~\cite{nao_ets},

\blockquote{ETS identified 97\% of all UK tests as “suspicious”. It classified 58\% of 58,459 UK tests as “invalid” and 39\% as “questionable”. The Home Office did not have the expertise to validate the results nor did it, at this stage, get an expert opinion on the quality of the voice recognition evidence. ... but the Home Office started cancelling visas of those individuals given an “invalid” test.}

The staggering number of accusations obviously included a number of false positives. The accuracy of ETS's method was disputed between experts sought by the National Union of Students and the Home Office; the resulting estimates of error rates ranged from 1\% to 30\%.
Yet out of 12,500 people who appealed their immigration decisions, only 3,600 won their cases---and only a fraction of these were won through actually disproving the allegations of cheating.
This highly opaque system was thus notable for the disproportionate amount of emphasis that was put into finding cheaters rather than protecting those who were falsely accused. Although we cannot be sure the voice recognition model was trained to optimize for sensitivity rather than specificity, as the head of the NAO aptly put, "When the Home Office acted vigorously to exclude individuals and shut down colleges involved in the English language test cheating scandal, we think they should have taken an equally vigorous approach to protecting those who did not cheat but who were still caught up in the process, however small a proportion they might be"~\cite{nao_ets}. This is an example of a system that was not designed to prevent a particular type of harmful failure.

\paragraph{Model Implementation Failures}

Even if a model was conceptualized in a reasonable way, some component of the system downstream from the original plan can be executed badly, lazily, or wrong. In 2011, the state of Idaho attempted to build an algorithm to set Medicaid assistance limits for individuals with developmental and intellectual disabilities. When individuals reported sudden drastic cuts to their allowances, the ACLU of Idaho tried to find out how the allowances were being calculated, only to be told it was a trade secret. The  class action lawsuit that followed resulted in a court-ordered disclosure of the algorithm, which was revealed to have critical flaws. According to Richard Eppink, Legal Director of the ACLU of Idaho~\cite{aclu_idaho},

\blockquote{
% We hired a couple of experts to dig into it and figure out what it was doing—how the whole process was working, both the assessment—the formula itself—and the data that was used to create it. 
There were a lot of things wrong with it.
First of all, the data they used to come up with their formula for setting people’s assistance limits was corrupt. They were using historical data to predict what was going to happen in the future. But they had to throw out two-thirds of the records they had before they came up with the formula because of data entry errors and data that didn’t make sense.}

Data validation is a critical step in the construction of a ML system, and the team that built the benefit system chose to use a highly problematic dataset to train their model. For this reason, we consider this to be an implementation failure.

Another way that failures can be attributed to poor implementation is when a testing framework was not appropriately implemented. 
%Failures can often be anticipated, and therefore addressed before deployment, through comprehensive testing.
% \lknote{Is "we tested it, but badly" an engineering or robustness failure?}
One area in which a lack of sufficient testing has been observed in the development of AI is in the area of clinical medicine. \citet{ai_vs_clinicians} systematically examined the methods and claims of studies which compared the performance of diagnostic deep learning computer vision algorithms against that of expert clinicians. In their literature review, they identified 10 randomized clinical trials and 81 non-randomized clinical trials. Of the 81 non-randomized studies, they found the median number of clinical experts compared to the AI was 4, full access to datasets and code were unavailable in over 90\% of studies, the overall risk of bias was high, and adherence to reporting standards were suboptimal, and therefore poorly substantiate their claims. %\drerror{Are these not post-deployment issues? Since we're talking about external validity here? Though I see why it's here and I'm not mad if we keep it here}
Similarly, the Epic sepsis prediction model, a product actually implemented at hundreds of hospitals, was recently externally validated by \citet{sepsis_validation}, who found that the model had poor calibration to other hospital settings and discriminated against under-represented demographics. %While it is unlikely that this model was not tested at all before deployment, 
These results suggest that the model's testing prior to deployment may have been insufficient to estimate its real-world performance.
Notably, the COVID-19 technology which resulted from innovation policy and democratization efforts mentioned in section \ref{assumption} 
%\lknote{fix ref}\drnote{done} 
was later shown to be completely unsuitable for clinical deployment after the fact~\cite{covidfail_summ, covidfail1, covidfail2, covidfail3}.

%\drerror{TO DO: Add Tesla/Uber}

%\drerror{Reference to section 3}
%In reality, the practical implementation of AI systems struggle to align with expectations at times simply because the declared objective or understood objective is not properly evaluated for, and the system fails in unarticulated ways. 
%The functionality assumption thus often leads to a fundamental confusion about what AI is, with the myth of how these algorithms operate superseding the reality of their technical details, flaws and limitations [cite]. This can lead to confusion about which problems pose the greatest threat to user safety. %As a result, ethical discussions focus disproportionately on regulating the malicious generation of fake content with these models~\cite{deepmind paper, openaI paper} and fail to acknowledge harms that arise from the more mundane hazard of the premature deployment of a system that simply does not work.
%%We see this assumption of model competence amplified in specific contexts. For example, many with ethical concerns surrounding large language models focus on hyped narratives of misinformation threats from generating hyper-realistic online content [cite], or the embedded moral reasoning of these models~\cite{delphi}. 

\paragraph{Missing Safety Features}

Sometimes model failures are anticipated yet difficult to prevent; in this case, engineers can sometimes take steps to ensure these points of failure will not cause harm. In 2014, a Nest Labs smoke and carbon monoxide detector was recalled~\cite{nest}. The detector had a feature which allowed the user to turn it off with a ``wave" gesture. However, the company discovered in testing that under certain circumstances, the sensor could be unintentionally deactivated. Detecting a wave gesture with complete accuracy is impossible, and Google acknowledges factors that contribute to the possibility of accidental wave triggering for its other home products~\cite{google_nest_help}. However, the lack of a failsafe to make sure the carbon monoxide detector could not be turned off accidentally made the product dangerous.

In the same way, the National Transportation Safety Board (NTSB) cited a lack of adequate safety measures---such as ``a warning/alert when the driver’s hands are off the steering wheel'', ``remote monitoring of vehicle operators'' and even the companies’ ``inadequate safety culture''---as the probable causes in at least two highly publicized fatal crashes of Uber ~\cite{uber-crash, uber-crash-2} and Tesla ~\cite{tesla-crash,tesla-crash-2} self-driving cars. As products in public beta-testing, this lack of functional safeguards was considered to be an even more serious operational hazard than any of the engineering failures involved (such as the vehicle's inability to detect an incoming pedestrian~\cite{uber-crash} or truck~\cite{tesla-crash}). 

This category also encompasses algorithmic decision systems in critical settings that lack a functional appeals process. This has been a recurring feature in algorithms which allocate benefits on behalf of the government~\cite{virginia_eubanks}. Not all of these automated systems rely on machine learning, but many have been plagued by bugs and faulty data, resulting in the denial of critical resources owed to citizens. In the case of the Idaho data-driven benefit allocation system, even the people responsible for reviewing appeals were unable to act as a failsafe for the algorithm’s mistakes: ``They would look at the system and say, `It’s beyond my authority and my expertise to question the quality of this result' ''~\cite{verge_aclu_idaho}.

\subsubsection{Deployment Failures}

Sometimes, despite attempts to anticipate failure modes during the design phase, the model does not ``fail'' until it is exposed to certain external factors and dynamics that arise after it is deployed.

\paragraph{Robustness Issues}

A well-documented source of failure is a lack of robustness to changing external conditions. \citet{LiaoAreWe2021} have observed that the benchmarking methods used for evaluation in machine learning can suffer from both internal and external validity problems, where “internal validity refers to issues that arise within the context of a single benchmark” and “external validity asks whether progress on a benchmark transfers to other problems.” If a model is developed in a certain context without strong evaluation methods for external validity, it may perform poorly when exposed to real-world conditions that were not captured by the original context. For instance, while many computer vision models developed on ImageNet are tested on synthetic image perturbations in an attempt to measure and improve robustness, but \citet{measuring_robustness_to_natural_distribution_shifts} have found that these models are not robust to real-world distribution shifts such as a change in lighting or pose. 

%\drerror{connect to section 3}
% over the years, as more and more AI tools found their way into healthcare~\cite{nature_survey}, it became clear that aspirations for such high stakes applications were largely overblown, with several tools and products falling short in independent audits, pilots and external validation tests~\cite{IBM, sepsis_epic_tool, mammography}. 

%\lknote{Are there more examples that illustrate this cleanly? I'm not sure about face mask NIST, chest x rays, etc} \drnote{why not? NIST mask is a robustness issue, though I guess that could also be seen as adversarial and there's not really any harm}

%\drerror{Todo: Rewrite/expand this paragraph to fit in this section}\drnote{Done}
Robustness issues are also of dangerous consequence in language models. 
For example, when large language models are used to process the queries of AI-powered web search \cite{nayak2019understanding}, the models' fragility to 
misspellings~\cite{moradi2021evaluating, pruthi2019combating}, or trivial changes to  format~\cite{bender2020climbing} and context~\cite{bender2021dangers} can lead to unexpected results. In one case, a large language model used in Google search could not adequately handle cases of  negation~\cite{ettinger2020bert} -- and so when queried with ``what to do when having a seizure'', the model alarmingly sourced the information for what \emph{not} to do, unable to differentiate between the two cases~\cite{goog_search_fail}. 
%The consequence of this failure was ultimately a false search result providing alarmingly false information to the user, 
%comprehend simple  
%where these problems range from failed negation comprehension~\cite{ettinger2020bert} to a poor handling of  to poorly scoped contexts for deployment~\cite{bender2021dangers}. Once deployed, these failures can distort in serious ways; for example, the results of AI-powered web search can lead to misinformation -- not due to the malicious use of a system functioning too well, but due to a system not functioning well enough~\cite{goog_search_fail}. \lkerror{Andrew marked this para with a ?. Care to elaborate? (Deb's paragraph, I haven't read this stuff.)} \aserror{Yeah, this paragraph is unintelligible to me.} 
%\drnote{It was a little... technically worded. I just re-wrote more simply.}

\paragraph{Failure under Adversarial Attacks}

Failures can also be induced by the actions of an adversary---an actor deliberately trying to make the model fail. Real-world examples of this often appear in the context of facial recognition, in which adversaries have some evidence that they can fool face-detection systems with, such as 3d-printed masks~\cite{3d_printed_masks} or software-generated makeup~\cite{makeup}. Machine learning researchers have studied what they call “adversarial examples,” or inputs that are designed to make a machine learning model fail~\cite{adversarial_examples}. However, some of this research has been criticized by its lack of a believable threat model— in other words, not focusing on what real-world “adversaries” are actually likely to do~\cite{catherine_olsson}.

\paragraph{Unanticipated Interactions}

A model can also fail to account for uses or interactions that it was not initially conceived to handle. %should have been anticipated. %Such a model does not work. 
%\lkerror{Not quite sure how to "Rephrase from maker perspective" as Andrew suggested. Anyone is free to try} 
Even if an external actor or user is not deliberately trying to break a model, their actions may induce failure if they interact with the model in a way that was not planned for by the model's designers. 
For instance, there is evidence that this happened at the Las Vegas Police Department:

\blockquote{As new records about one popular police facial recognition system show, the quality of the probe image dramatically affects the likelihood that the system will return probable matches. But that doesn't mean police don't use bad pictures anyway. According to documents obtained by Motherboard, the Las Vegas Metropolitan Police Department (LVMPD) used ``non-suitable" probe images in almost half of all the facial recognition searches it made last year, greatly increasing the chances the system would falsely identify suspects, facial recognition researchers said.~\cite{vegas_pd}}

%This is still a failure of the technology itself even if it was induced by external factors. In particular, i
This aligns with reports from ~\citet{garvie2019garbage} about other police departments inappropriately uploading sketch and celebrity photos to facial recognition tools. It is possible for designers to preempt misuse by implementing instructions, warnings, or error conditions, and failure to do so creates a system that does not function properly.

\subsubsection{Communication Failures}

%% Sometimes its abou\t ommission of information
%\drnote{I really love this intro!}
As with other areas of software development, roles in AI development and deployment are becoming more specialized. Some roles focus on managing the data that feeds into models, others specialize in modeling, and others optimally engineer models for speed and scale~\cite{De_Mauro2018-mi}. There are even those in "analytics translator" roles -- managers dedicated to acting as communicators between data science work and non-technical business leaders~\cite{Henke2018-ua}. And, of course, there are salespeople. Throughout this chain of actors, potential miscommunications or outright lies can happen about the performance, functional safety or other aspects of deployed %or for sale 
AI/ML systems. Communication failures often co-occur with other functional safety problems, and the lack of accountability for false claims -- intentional or otherwise -- makes these particularly pernicious and likely to occur as AI hype continues absent effective regulation. 

\paragraph{Falsified or Overstated Capabilities}

To pursue commercial or reputational interests, companies and researchers may explicitly make claims about models which are provably untrue. A common form of this are claims that a product is "AI", when in fact it mainly involves %the product is 
humans making decisions behind the scenes.
While this in and of itself may not create unsafe products, expectations based on unreasonable claims can create unearned trust, and a potential over-reliance that hurts parties who purchase the product. As an example, investors poured money into ScaleFactor, a startup that claimed to have AI that could replace accountants for small businesses, with the exciting (for accountants) tagline "Because evenings are for families, not finance"~\cite{scalefactor}. Under the hood, however, 

\blockquote{Instead of software producing financial statements, dozens of accountants did most of it manually from ScaleFactor’s Austin headquarters or from an outsourcing office in the Philippines, according to former employees. Some customers say they received books filled with errors, and were forced to re-hire accountants, or clean up the mess themselves.~\cite{scalefactor}
}% \asnote{Also, worth making a quick reference to the OG mechanical turk? The fake robot with the guy playing chess hidden inside? Not necessary, but made me think of it.} \lknote{I have a turkish friend who takes offense at this metaphor LOL}

%% Note I am wondering if this belongs in misconstrued capabilities as it seems to be a form of metrics gaming
Even large well-funded entities misrepresent the capabilities of their AI products. Deceptively constructed evaluation schemes allow AI product creators to make false claims. In 2018, Microsoft created machine translation with "equal accuracy to humans in Chinese to English translations"~\cite{Translator2018-ki}. However, the study used to make this claim (still prominently displayed in press release materials) was quickly debunked by a series of outside researchers who found that at the document-level, when provided with context from nearby sentences, and/or compared to human experts, the machine translation model did not indeed achieve equal accuracy to human %s in Chinese to English
translators~\cite{Toral2018-wn,Laubli2018-sn}. This follows a pattern seen with machine learning products in general, where the advertised performance on a simple and static data benchmark, is much lower than the performance on the often more complex and diverse data encountered in practice. % which that often    \drerror{reporting performance on a test set vs. realistic setting}

\paragraph{Misrepresented Capabilities}
A simple way to deceive customers into using prediction services is to sell the product for a purpose you know it can't reliably be used for. In 2018, the ACLU of Northern California revealed that Amazon effectively misrepresented capabilities to police departments in selling their facial recognition product, Rekognition. Building on previous work~\cite{Buolamwini_undated-dd}, the ACLU ran Rekognition with a database of mugshots against members of U.S. Congress using the default setting and found 28 members falsely matched within the database, with people of color shown as a disproportionate share of these errors~\cite{Snow2018-vw}. This result was echoed by \citet{raji2019actionable} months later. Amazon responded by claiming that for police use cases, the threshold for the service should be set at either 95\% or 99\% confidence %-- it is difficult to tell based on the choices made in evaluation
~\cite{Wood_undated-ek}. However, based on a detailed timeline of events~\cite{aclu_response_response_fr}, it is clear that in selling the service through blog posts and other campaigns that thresholds were set at 80\% or 85\% confidence, as the ACLU had used in its investigation. In fact, suggestions to shift that threshold were buried in manuals end-users did not read or use -- even when working in partnership with Amazon. At least one of Amazon's police clients also claimed being unaware of needing to modify the default threshold~\cite{menegus2019defense}. 

The hype surrounding IBM's Watson in healthcare represents another example where a product that may have been fully capable of performing \emph{specific} helpful tasks was sold as a panacea to health care's ills. As discussed earlier, this is partially the result of functional failures like practical impossibility -- but these failures were coupled with deceptively exaggerated claims. The backlash to this hype has been swift in recent years, with one venture capitalist claiming "I think what IBM is excellent at is using their sales and marketing infrastructure to convince people who have asymmetrically less knowledge to pay for something"~\cite{Wojcik_undated-nb}. At Memorial-Sloan Kettering, after \$62 million dollars spent and may years of effort, MD Anderson famously cancelled IBM Watson contracts with no results to show for it~\cite{md_anderson_benches_watson}. %Common functional failures coupled with exaggerated claims can create trust that is not deserved, lead to public spend on dubious efforts, and create opportunity costs through halting progress that could actually help people.

%\drnote{re-written}
%\citet{virginia_eubanks} and \citet{green2019smart} reveal how
This is particularly a problem in the context of algorithms developed by public agencies -- where the AI systems can be adopted
%almost disingenuously, introduced as
as symbols for progress, or smokescreens for undesirable policy outcomes, and thus liable to inflated narratives of performance. \citet{green2019smart} discusses how the celebrated success of ``self-driving shuttles'' in Columbus, Ohio omits its marked failure in the lower-income Linden neighborhood, where residents were now locked out of the transportation apps due to a lack of access to a bank account, credit cards, a data plan or Wi-Fi. Similarly, \citet{virginia_eubanks} demonstrates how a \$1.4 billion contract with a coalition of high-tech companies led an Indiana governor to stubbornly continue a welfare automation algorithm that resulted in a 54\% increase in the denials of welfare applications. %and how similar city investments disguised the failures of Allegeny County's child welfare case triaging tool and the Los Angeles 

%locked out of transportation apps because they lack access to a bank account or credit card
%lack access to a data plan or Wi-Fi

%or initiatives of Sidewalk Labs in New York City or Toronto, Canada 

%For example, a 

%, such programs tend to not be effective at solving the problem at hand, but are mis-represented as functional and deployed anyways, to achieve these political ends.

%% threshold in hiring algorithms if I can find it publically ; in order to pass EEOC standards, companies will run validations to claim no aderse impact but they do this by shifting the threshold such that they meet the criteria, even if real use case cant possibly work at that threshold. I'm thinking maybe we avoid as i dont want to make this about bias

%% Lying by Omission 

%% police robot giving vibes it would be helpful
% https://www.nbcnews.com/tech/tech-news/robocop-park-fight-how-expectations-about-robots-are-clashing-reality-n1059671

%% \paragraph{Misrepresentation by Omission}

%% \aherror{Todo - Decide whether this is sufficiently different / usable from other errors}

%% Examples of situations where product underreports evaluation -- doesnt give enough details to reproduce it

%% Arvind -Irreproducibly in Machine Learning
%% Metareview of clinical studies found all scientific valuation studies werent supported by methods

%% Withholding Information Needed to Properly Evaluate
%% paper about algorithm hiring tools, but they reveal that none of tools had published validation process (misrepresentation through omission); hirevue performance ; 

\section{Dealing With Dysfunction: Opportunities for Intervention on Functional Safety}
\label{responses}
%Part II: Legal Levers/ Policy discussion [Deb + Andrew] - 2-3 + table 
%“Legal review” of current AI Accountability policy, discussing missing links 
%Current assumptions about functionality 
%Future policy directions / relevant legal frameworks

%Assumptions of Functionality
%Another example comes from...

%New AI policy proposals often embed a similar assumption. 
%\subsection{Modes of Intervention}
%Considerations 
%Aspirations Intervention types 

%FDA
%Three opportunities to interfere

%Intro to this section 
%Market entry, advertising and promotion [during] and post-market surveillance 

%\subsection{Assumptions of Functionality}

%To a large extent, policymakers and scholarly communities have focused their discussion, critiques, and proposed reforms with the background assumption that AI basically works as advertised. But the assumption has also left out many levers that already exist and deal with products that just don't work.

The challenge of dealing with an influx of fraudulent or dysfunctional products is one that has plagued many industries, including food safety~\cite{blum2018poison}, medicine~\cite{anderson2015snake,bausell2009snake}, financial modeling~\cite{silver2012signal}, civil aviation~\cite{heppenheimer1995turbulent} and the automobile industry~\cite{nader1965unsafe,vinsel2019moving}. %, in addition to others.
In many cases, it required the active advocacy of concerned citizens %- figures like Ralph Nader for the automobile Industry, Harvey Washington Wiley for food safety, and Samuel Hopkins Adams for medicine, among others - to counteract the financial and political incentives that led these industries to disguise the failed effectiveness of popular products~\cite{nader1965unsafe, blum2018poison}. This advocacy fed directly into 
to lead to the policy interventions that would effectively change the tide of these industries. The AI field seems to now be facing this same challenge. %, and may require similar organizational and policy interve %- developing in the passage of landmark legislation (such as the Pure Food and Drug Act,  the Air Commerce Act, Consumer Product Safety Act, and National Traffic and Motor Vehicle Safety Act), some of which would eventually evolve into federal enforcement agencies (such as the Federal Food and Drug Administration (FDA), Federal Aviation Administration (FAA), and National Highway Traffic Safety Administration(NHTSA)). 

Thankfully, as AI operates as a general purpose technology prevalent in many of these industries, there already exists a plethora of governance infrastructure to address this issue in related fields of application. %that indicate the critical role functionality plays in protecting downstream populations from harm. The visibility of current AI failures today is often tied to the fact that the technology is currently being deployed in contexts that already have regulatory and organizational measures in place, that have begun to catch these failures.  
In fact, healthcare is the field where AI product failures appear to be the most visible, in part due to the rigor of pre-established evaluation processes~\cite{wu2021medical,liu2020reporting, rivera2020guidelines, benjamens2020state}. %, FDA_process_ref, AI_clinical_trail_ref, FDA_approved_algo}. %Healthcare regulation, on the pharmaceutical side also has a rich history of post-market surveillance, ie. monitoring deployed products after market entry for adverse effects. This also makes failures and their consequences easier to identify, and constrains market-viable products to those that work. 
Similarly, the transportation industry has a rich history of thorough accident reports and investigations, through organizations such as the National Transportation and Safety Board (NTSB), who have already been responsible for assessing the damage from the few known cases of self-driving car crashes from Uber and Tesla~\cite{harris2019ntsb}.

In this section, we specifically outline the legal and organizational interventions necessary to address functionality issues in general context in which AI is developed and deployed into the market. In broader terms, the concept of \emph{functional safety} in engineering design literature~\cite{smith2004functional,roland1991system} well encapsulates the concerns articulated in this paper---namely that a system can be deployed without working very well, and that such performance issues can cause harm worth preventing. 

%\drerror{might move into an appendix table?}
%Also, note that there are several considerations to keep in mind when classifying types of failures and reflecting on the connection between functional failure and accountability. These considerations include notions of \emph{intentionality} (ie. What were the AI developers’ intentions? Was it an accident (no intention to fail) or sabotage (intention to fail)?), \emph{blameworthiness} (ie. How much is the AI developer responsible for? How much did they do relative to known public expectations or legal liability obligations?), \emph{awareness} (ie. How much was known about this failure mode?), \emph{observability} (ie. How visible is this failure mode?), \emph{foreseeability} (ie. Which failures were expected to be visible or known?), \emph{reasonableness} (ie. How much did AI developers act according to expectations?) and \emph{substantiation} (ie. how much evidence is there to validate the AI system?). 

\subsection{Legal/Policy Interventions}
%\drnote{Do we need to mention negligence tort & functionality as a basic requirement somewhere?}
% hey look out for these things, there are existing policy claims that are compatible 

The law has several tools at its disposal to address product failures to work correctly. They mostly fall in the category of consumer protection law. This discussion will be U.S.-based, but analogues exist in most jurisdictions.

\subsubsection{Consumer Protection}
The Federal Trade Commission is the federal consumer protection agency within the United States with the broadest subject matter jurisdiction. Under Section 5 of the FTC Act, it has the authority to regulate ``unfair and deceptive acts or practices'' in commerce~\cite{Section_5}. This is a  broad grant authority to regulate practices that injure consumers. The authority to regulate deceptive practices applies to any material misleading claims relating to a consumer product. The FTC need not show intent to deceive or that deception actually occurred, only that claims are misleading. Deceptive claims can be expressed explicitly---for example, representation in the sales materials that is inaccurate---or implied, such as an aspect of the design that suggests a functionality the product lacks~\cite{hoofnagle2016federal,hartzog2018privacy}. Many of the different failures, especially impossibility, can trigger a deceptive practices claim.

The FTC's ability to address unfair practices is wider-ranging but more controversial. The FTC can reach any practice ``likely to cause substantial injury to consumers[,] not reasonably avoidable by consumers themselves and not outweighed by countervailing benefits to consumers'' ~\cite{Section_5}. Thus, where dysfunctional AI is being sold and its failures causes substantial harm to consumers, the FTC could step in. Based on the FTC's approach to data security, in which the Commission has sued companies for failing to adequately secure consumer data in their possession against unknown third-party attackers~\cite{mcgeveran2018duty}, even post-deployment failures---if foreseeable and harmful---can be included among unfair practices, though they partially attributable to external actors.

The FTC can use this authority to seek an injunction, requiring companies to cease the practice. Formally, the FTC does not have the power to issue fines under its Section 5 authority, but the Commission frequently enters into long-term consent decrees with companies that it sues, permitting continuing jurisdiction, monitoring, and fines for future violations~\cite{FB_consent_decree, Snapchat_consent_decree}. The Commission does not have general rulemaking authority, so most of its actions to date have taken the form of public education and enforcement. The Commission does, however, have authority to make rules regarding unfair or deceptive practices under the Magnuson-Moss Warranty Act. Though it has created no new rules since 1980, in July 2021, the FTC voted to change internal agency policies to make it easier to do so~\cite{FTC_Mag_Moss}.

Other federal agencies also have the ability to regulate faulty AI systems, depending on their subject matter. The Consumer Product Safety Commission governs the risks of physical injury due to consumer products. They can create mandatory standards for products, can require certifications of adherence to those rules, and can investigate products that have caused harm, leading to bans or mandatory recalls ~\cite{CPSC_about}. The National Highway Safety Administration offers similar oversight for automobiles specifically. The Consumer Finance Protection Bureau can regulate harms from products dealing with loans, banking, or other consumer finance issues ~\cite{CFPB_jx}. 

In addition to various federal agencies, all states have consumer protection statutes that bar deceptive practices and many bar unfair practices as well, like the FTC Act~\cite{NCLC_Report}. False advertising laws are related and also common. State attorneys general often take active roles as enforcers of those laws~\cite{citron2016privacy}. Of course, the efficacy of such laws varies from state to state, but in principle, they become another source of law and enforcement to look to for the same reasons that the FTC can regulate under Section 5. One particular state law worth noting is California's Unfair Competition Law, which allows individuals to sue for injunctive relief to halt conduct that violates other laws, even if individuals could not otherwise sue under that law ~\cite{Zhang}.

It is certainly no great revelation that federal and state regulatory apparatuses exist. Rather, our point is that while concerns about discrimination and due process can lead to difficult questions about the operation of existing law and proposals for legal reform, thinking about the ways that AI is \emph{not working} makes it look like other product failures that we know how to address. Where AI doesn't work, suddenly regulatory authority is easy to find.

\subsubsection{Products Liability Law}

Another avenue for legal accountability may come from the tort of products liability, though there are some potential hurdles. In general, if a person is injured by a defective product, they can sue the producer or seller in products liability. The plaintiff need not have purchased or used the product; it is enough that they were injured by it, and the product has a defect that rendered it unsafe.

%\aserror{Discuss product misuse}
%The heart of products liability law is in the definition of defective products. With the caveat that we are speaking about U.S. law, and can only speak generally, as each state has variations to its law, defects come in one of three types. A manufacturing defect occurs when a product differs from its intended design in a way that is unreasonably dangerous~\cite{ThirdRestatement}. A design defect occurs where "the foreseeable risks of harm posed by the product could have been reduced or avoided by the adoption of a reasonable alternative design."~\cite{ThirdRestatement} This is essentially a cost-benefit test. A warning defect occurs where foreseeable harms from a product could have been "reduced or avoided by the provision of reasonable instructions or warnings."~\cite{ThirdRestatement}. If the product in question contains one of these defects, and such defect caused the plaintiff's injury, the defendant is liable for the injury. 

It would stand to reason that a functionality failure in an AI system could be deemed a product defect. But surprisingly, defective software has never led to a products liability verdict. One commonly cited reason is that products liability applies most clearly to tangible things, rather than information products, and that aside from a stray comment in one appellate case~\cite{Winter}, no court has actually ruled that software is even a ``product'' for these purposes~\cite{calo2015robotics, engstrom20133D}. This would likely not be a problem for software that resides within a physical system, but for non-embodied AI, it might pose a hurdle. In a similar vein, because most software harms have typically been economic in nature, with, for example, a software crash leading to a loss of work product, courts have rejected these claims as "pure economic loss" belonging more properly in contract law than tort. But these mostly reflect courts' anxiety with intangible \textit{injuries}, and as AI discourse has come to recognize many concrete harms, these concerns are less likely to be hurdles going forward~\cite{choi2019crashworthy}.

Writing about software and tort law, \citet{choi2019crashworthy} identifies the complexity of software as a more fundamental type of hurdle. For software of nontrivial complexity, it is provably impossible to guarantee bug-free code. %, so courts will be loathe to impose liability on code that simply behaves incorrectly. 
An important part of products liability is weighing the cost of improvements and more testing against the harms. But as no amount of testing can guarantee bug-free software, it will difficult to determine how much testing is enough to be considered reasonable or non-negligent ~\cite{choi2019crashworthy, hubbard2014sophisticated}. Choi analogizes this issue to car crashes: car crashes are inevitable, but courts developed the idea of crashworthiness to ask about the car's contribution to the total harm, even if the initial injury was attributable to a product defect~\cite{choi2019crashworthy}. While Choi looks to crashworthiness as a solution, the thrust of his argument is that software can cause exactly the type of injury that products liability aims to protect us from, and doctrine should reflect that. %If courts follow that suggestion, software relative's robustness could become a measure of liability.

While algorithmic systems have a similar sort of problem, the failure we describe here are more basic. Much as writing bug-free software is impossible, creating a model that handles every corner case perfectly is impossible. But the failures we address here are not about unforeseeable corner cases in models. We are concerned with easier questions of basic functionality, without which a system should never have been shipped. If a system is not functional, in the sense we describe, a court should have no problem finding that it is unreasonably defective.
%Despite being software, however, AI's failures as we describe them are not a direct analogue to crashes. The proper AI analogy to the software crash problem would be a situation where AI fundamentally works as advertised, but runs into unanticipated corner cases. In a ML implementation, there will always be corner cases that are not predicted, and continued testing may or may not help find them~\cite{selbst2020negligence}. That is therefore a direct analogue to the impossibility of finding all software crashes that makes product liability complicated. The errors we describe here are more fundamental.
As discussed above, a product could be placed on the market claiming the ability to do something it cannot achieve in theory or in practice, or it can fail to be robust to unanticipated but foreseeable uses by consumers. Even where these errors might be difficult to classify in doctrinally rigid categories of defect, courts have increasingly been relying on ``malfunction doctrine,'' which allows for circumstantial evidence to be used as proof of defect where ``a product fails to perform its manifestly intended function.''~\cite{ThirdRestatement_S3}. Courts are increasingly relying on this doctrine and it could apply here~\cite{owen2001manufacturing, geistfeld2017roadmap}. Products liability could especially easily apply to engineering failures, where the error was foreseeable and an alternative, working version of the product should have been built.

\subsubsection{Warranties}

Another area of law implicated by product failure is warranty law, which protects the purchasers of defunct AI and certain third parties who stand to benefit from the sale. Sales of goods typically come with a set of implied warranties. The implied warranty of merchantability applies to all goods and states, among other things, that the good is ``fit for the ordinary purposes for which such goods are used''~\cite{UCC_2-314}. The implied warranty of fitness for particular purpose applies when a seller knows that the buyer has a specific purpose in mind and the buyer is relying on the seller's skill or judgment about the good's fitness, stating that the good is fit for that purpose\cite{UCC_2-315}. Defunct AI will breach both these warranties. 
The remedy for such a breach is limited to contract damages. This area of law is concerned with ensuring that purchasers get what they pay for, so compensation will be limited roughly to value of the sale. Injuries not related to the breach of contract are meant to be worked out in tort law, as described above.

\subsubsection{Fraud}

In extreme cases, the sale of defunct AI may constitute fraud. Fraud has many specific meanings in law, but invariably it involves a knowing or intentional misrepresentation that the victim relied on in good faith. In contract law, proving that a person was defrauded can lead to contract damages. Restitution is another possible remedy for fraud. In tort law, a claim of fraud can lead to compensation necessary to rectify any harms that come from the fraud, as well as punitive damages in egregious cases. Fraud is difficult to prove, and our examples do not clearly indicate fraud, but it is theoretically possible if someone is selling snake oil. Fraud can lead to criminal liability as well.

\subsubsection{Other Legal Avenues Already Being Explored}

Finally, other areas of law that are already involved in the accountability discussion, such as discrimination and due process, become much easier cases to make when the AI doesn't work. Disparate impact law requires that the AI tool used be adequately predictive of the desired outcome, before even getting into the question of whether it is \textit{too} discriminatory or not~\cite{barocas2016big}. A lack of construct validity would easily subject a model's user to liability. Due process requires decisions to not be arbitrary, and AI that doesn't work loses its claim to making decisions on a sound basis~\cite{citron2007technological}. Where AI doesn't work, legal cases in general become easier.

\subsection{Organizational interventions}
%[Deb] Look back at our taxonomy, make sure you don't do that 

%\subsection{Organizational interventions}
%\drerror{This is a first attempt at this -- wonder if I’m even approaching it correctly}
%\aherror{I think we need to point out here that org levers will become more useful/used in the face of regulation expectations}

In addition to legal levers, there are many organizational interventions that can be deployed to address the range of functionality issues discussed. %In this section, we break down what can be done by organizations to address functionality challenges. 
Due to clear conflicts of interest, the self-regulatory approaches described are far from adequate oversight for these challenges, and the presence of regulation does a lot to incentivise organizations to take these actions in the first place. However, they do provide an immediate path forward in addressing these issues. 

\subsubsection{Internal Audits \& Documentation}

%One of the most direct approaches to addressing functionality issues in organizations is by recording and maintaining thorough internal audit and engineering evaluation processes. %as part of the engineering development lifecycle of the AI product. 

After similar crises of performance in fields such as aerospace, finance and medicine, such processes evolved in those industries to enforce a new level of introspection in the form of internal audits. Taking the form of anything from documentation exercises to challenge datasets as benchmarks, these processes raised the bar for deployment criteria and matured the product development pipeline in the process~\cite{raji2020closing}. The AI field could certainly adopt similar techniques for increasing the scrutiny of their systems, especially given the nascent state of reflection and standardization common in ML evaluation processes~\cite{LiaoAreWe2021}. For example, the ``Failure modes, effects, and diagnostic analysis (FMEDA)’’ documentation process from the aerospace industry could support the identification of functional safety issues prior to AI deployment~\cite{raji2020closing}, in addition to other resources from aerospace (such as the functional hazard analyses (FHA) or Functional Design Assurance Levels (FDALS)). 

%Procedural standard for what activities need to be executed as part of the documentation process in addition to reporting standards and expectations are lacking in the AI field.

Ultimately, internal audits are a self-regulatory approach---though audits conducted by independent second parties such as a consultancy firm could provide a fresh perspective on quality control and performance in reference to articulated organizational expectations~\cite{UNESCO}. The challenge with such audits, however, is that the results are rarely communicated externally and disclosure is not mandatory, nor is it incentivized. As a result, assessment outcomes are mainly for internal use only, often just to set internal quality assurance standards for deployment and prompt further engineering reflection during the evaluation process. 

\subsubsection{Product Certification \& Standards}

A trickier intervention is the avenue of product certification and standards development for AI products. 
This concept has already made its way into AI policy discourse; CEN (European Committee for Standardisation) and CENELEC (European Committee for Electrotechnical Standardisation), two of three European Standardisation Organisations (ESOs) were heavily involved in the creation of the EU's draft AI Act~\cite{veale_euact}. On the U.S. front, industry groups IEEE and ISO regularly shape conversations, with IEEE going so far as to attempt the development of a certification program~\cite{ieee_cert, havens2019principles}. 
In the aviation industry, much of the establishment of engineering standards happened without active government intervention, between industry peers~\cite{raji2020closing}. These efforts resemble the Partnership on AI’s attempt to establish norms on model documentation processes~\cite{raji2019ml}. Collective industry-wide decision-making on critical issues can raise the bar for the entire industry and raise awareness within the industry of the importance of handling functionality challenges. 
Existing functional safety standards from the automobile (ISO 26262), aerospace (US RTCA DO-178C), defense (MIL-STD-882E) and electronics (IEEE IEC 61508 / IEC 61511) industries, amongst others, can provide a template on how to approach this challenge within the AI industry. 

%\subsubsection{Product Procurement Processes}

\subsubsection{Other Interventions}
%\drerror{Add citations}

There are several other organizational factors that can determine and assess the functional safety of a system. As a client making decisions on which projects to select, or permit for purchase, it can be good to set performance related requirements for procurement and leverage this procurement process in order to set expectations for functionality~\cite{sloane2021ai,mulligan2019procurement,richardson2021best,rubenstein2021acquiring}. Similarly, cultural expectations for safety and engineering responsibility impact the quality of the output from the product development process -- setting these expectations internally and fostering a healthy safety culture can increase cooperation on other industry-wide and organizational measures~\cite{roland1991system}. Also, as functionality is a safety risk aligned with profit-oriented goals, many model logging and evaluation operations tools are available for organizations to leverage in the internal inspection of their systems -- including tools for more continuous monitoring of deployed systems~\cite{ratner2019mlsys,shankar2021towards}.

\section{Conclusion : The Road Ahead}
%\drerror{TO DO to add: functionality is not completely ignored, but just not adequately emphasized}
%Many deployed AI systems don’t work. More often than not, policymakers need to start from this assumption rather than ignore this reality. Functionality issues are the neglected low-hanging fruit of AI regulation. At the moment, those developing these systems face no consequences for fraudulent or irresponsible behavior and there is little to no oversight in what AI-branded products have an obligation to do once put out on the market. 
We cannot take for granted that %policymakers understand when and how 
AI products work. 
%When regulators are beholden to corporate, academic and media guidance to dictate for them how well these systems function, they tend to all too often 
Buying into the presented narrative of a product with at least basic utility or an industry that will soon enough ``inevitably’’ overcome known functional issues causes us to miss important sources of harm and available legal and organizational remedies.
%In this paper, we demonstrate that in reality, this is not a valid assumption to hold. When we adopt this default view, we fail to address the serious harms that arise in consequence of dysfunction. 
Although functionality issues are not completely ignored in AI policy, the lack of awareness of the range in which these issues arise leads to the problems being inadequately emphasized and poorly addressed by the full scope of accountability tools available. 

The fact that faulty AI products are on the market today makes this problem particularly urgent. Poorly vetted products permeate our lives, and while many readily accept the potential for harms as a tradeoff, the claims of the products' benefits go unchallenged. But addressing functionality involves more than calling out demonstrably broken products. It also means challenging those who develop AI systems to better and more honestly understand, explore, and articulate the limits of their products prior to their release into the market or public use. %At the very least, any vetting process or evaluation measure put in place to verify claims of functionality provides an initial friction necessary to slowing down the entry of the product to the market, effectively blocking its way into the lives of those vulnerable to it’s impact. For products already on the market, once removed, entire product categories become more scrutinized than ever, facing even more hurdles for market re-entry. Such necessary pauses in the process provides an opportunity for contemplation and further assessment, creating the avenue for additional consideration on even more ethical concerns. 
Adequate assessment and communication of functionality should be a minimum requirement for mass deployment of algorithmic systems. Products that do not function should not have the opportunity to affect people's lives. %Functionality considerations are essential and  foundational step towards addressing overall safety and ethical concerns in the AI industry. 

%National Institute of Health’s Office of Alternative Medicine (OAM) 

%``The purpose of the OAM, I began to realize, was to demonstrate that these disparate therapies all work.''

%CITE: Voodoo Science, Robert Park 

%Pre-market approval, post-market surveillance and audits 

%Those over-concerned with AGI, singularity do that to deflect from serious conversations about functionality.
%Broader question for us : Do we have any examples where we think AI does work?!

%There are some narrow cases - genetic testing for example where AI has been useful.
%\drnote{call to action: our audience is to policy researchers - this is what you could be working on, knowing that this doesn't work}
%\drnote{future work or limitation: who is responsible and stakeholder ; we are tlaking about functionality and we know that functionality is not the only thing and ultimate responsibility is distributed}
%\drnote{usually it means the manufacturer is responsible; but their responsibility is limited by errors we can't see, and are made more responsible by unanticipated by things that can't happen}

\begin{acks}
We thank the Mozilla Foundation and the Algorithmic Justice League for providing financial support during this project. 
%Thnks 4 the moneys

%Can/should we thank Suresh here, since he's off the paper?
\end{acks}

\bibliographystyle{ACM-Reference-Format}
\bibliography{bib.bib}

%% \printbibliography

\end{document}